\documentclass[symmetry,article,accept,moreauthors,pdftex]{Definitions/mdpi}

\setitemize{parsep=6pt,itemsep=0pt,leftmargin=*,labelsep=5.5mm}
\setenumerate{parsep=6pt,itemsep=0pt,leftmargin=*,labelsep=5.5mm}
\setlist[description]{itemsep=0mm}

\firstpage{1} 
\pubvolume{xx}
\issuenum{1}
\articlenumber{5}
\pubyear{2019}
\copyrightyear{2019}
\history{Received: 11 September 2019; Accepted: 15 October 2019; Published: date}

\usepackage{graphicx} 
\usepackage[caption=false,font=footnotesize]{subfig}

\usepackage[ruled,vlined]{algorithm2e}
\usepackage{setspace}
\usepackage{xspace}

\usepackage{easybmat}
\usepackage{color}
\usepackage{enumitem}
\usepackage{cancel}

\newcommand{\mathds}[1]{\mathbb{#1}}
\newcommand{\beq}{\begin{equation}}
\newcommand{\eeq}{\end{equation}}
\newcommand{\R}{\mathds{R}}
\newcommand{\E}{\mathds{E}}

\newcommand{\Prob}{\mathds{P}}

\newcommand{\cK}{{\mathcal K}}

\newcommand{\cN}{{\mathcal N}}
\newcommand{\cO}{{\mathcal O}}

\newcommand{\vlam}{{\lambda}}

\newcommand{\tr}{\text{tr}}

\DeclareMathOperator*{\argmin}{arg\,min}

\newcommand{\prox}{\textnormal{prox}}

\newcommand{\gslope}{nsSLOPE\xspace}
\newcommand{\glasso}{gLASSO\xspace}
\newcommand{\rx}{\mathrm{x}}

\usepackage{bbm}
\newcommand{\vone}{\mathbbm 1}

\Title{Structure Learning of Gaussian Markov Random Fields with False Discovery Rate Control}

\Author{Sangkyun Lee $^{1}$\orcidA{}*, Piotr Sobczyk $^{2}$ and Malgorzata Bogdan $^{3}$}

\AuthorNames{Sangkyun Lee, Piotr Sobczyk and Malgorzata Bogdan}

\address{%
  $^{1}$ \quad  Computer Science, Hanyang University ERICA, Ansan 15588, Korea\\
 $^{2}$ \quad Department of Mathematics, Wroclaw University of Science and Technology, 50-370 Wroclaw, Poland; piotr.sobczyk@pwr.edu.pl\\
 $^{3}$ \quad Institute of Mathematics, University of Wroclaw, 50-384 Wroclaw, Poland; Malgorzata.Bogdan@uwr.edu.pl\\
}

\corres{Correspondence: sangkyun@hanyang.ac.kr}

\abstract{ In this paper, we propose a new estimation procedure for
  discovering the structure of Gaussian Markov random fields (MRFs)
  with false discovery rate (FDR) control, making use of the sorted
  $\ell_1$-norm (SL1) regularization. A Gaussian MRF is an acyclic
  graph representing a multivariate Gaussian distribution, where nodes
  are random variables and edges represent the conditional dependence
  between the connected nodes. Since it is possible to learn the edge
  structure of Gaussian MRFs directly from data, Gaussian MRFs provide
  an excellent way to understand complex data by revealing the
  dependence structure among many inputs features, such as genes,
  sensors, users, documents, etc. In learning the graphical structure
  of Gaussian MRFs, it is desired to discover the actual edges of the
  underlying but unknown probabilistic graphical model—it becomes
  more complicated when the number of random variables (features) $p$
  increases, compared to the number of data points $n$. In particular,
  when $p \gg n$, it is statistically unavoidable for any estimation
  procedure to include false edges. Therefore, there have been many
  trials to reduce the false detection of edges, in particular, using
  different types of regularization on the learning parameters. Our~method makes use of the SL1 regularization, introduced recently for
  model selection in linear regression. We~focus on the benefit of SL1
  regularization that it can be used to control the FDR of detecting
  important random variables. Adapting SL1 for probabilistic graphical
  models, we show that SL1 can be used for the structure learning of
  Gaussian MRFs using our suggested procedure \gslope (neighborhood selection Sorted
  L-One Penalized Estimation), controlling the FDR of detecting
  edges.}


\begin{document}

\section{Introduction}

Estimation of the graphical structure of Gaussian Markov random fields
(MRFs) has been the topic of active research in machine learning, data
analysis and statistics. The reason is that they provide efficient
means for representing complex statistical relations of many variables
in forms of a simple undirected graph, disclosing new insights about
interactions of genes, users, news articles, operational~parts of a
human driver, to name a few.

One mainstream of the research is to estimate the structure by maximum
likelihood estimation (MLE), penalizing the $\ell_1$-norm of the
learning parameters. In this framework, structure learning of a
Gaussian MRF is equivalent to finding a sparse inverse covariance
matrix of a multivariate Gaussian distribution.  To formally describe
the connection, let us we consider $n$ samples $x_1,x_2,\dots,x_n$ of
$p$ jointly Gaussian random variables following
$\cN(0, \Sigma_{p\times p})$, where the mean is zero without loss of
generality and $\Sigma_{p\times p}$ is the covariance matrix. The
estimation is essentially the MLE of the inverse covariance matrix
$\Theta := \Sigma^{-1}$ under an $\ell_1$-norm penalty, which can be
stated it as a convex optimization problem~\cite{YuaL07}:
\begin{equation}\label{eq:glasso}
 \hat\Theta = \argmin_{\Theta \in \R^{p\times p}, \Theta'=\Theta} \; -\log\det \Theta + \tr(S\Theta) + \lambda \|\Theta\|_1
\end{equation}

Here, $S := \frac{1}{n} \sum_{i=1}^n x_i x_i'$ is the sample
covariance matrix, $\|\Theta\|_1 := \sum_{i,j} |\Theta_{ij}|$ and
$\lambda>0$ is a tuning parameter that determines the element-wise
sparsity of $\hat\Theta$.

The $\ell_1$-regularized MLE approach~\eqref{eq:glasso} has been
addressed quite extensively in literature. The~convex optimization
formulation has been first discussed in References \cite{AspB08, BanG08}. A
block-coordinate descent type algorithm was developed in Reference
\cite{FriH08}, while revealing the fact that the sub-problems of
\eqref{eq:glasso} can be solved in forms of the LASSO regression
problems~\cite{Tib96}. More efficient solvers have been developed to
deal with high-dimensional cases~\cite{OztN12, RolB12, HsiD11, HsiB12,
  HsiS13, MazH12, TreT14, ZhaF18}. In the theoretical side, we are
interested in two aspects—one is the estimation quality of
$\hat\Theta$ and another is the variable selection quality of
$\hat\Theta$. These aspects are still in under investigation in theory
and
experiments~\cite{MeiB06,MeiB10,YuaL07,BanG08,RotB08,LamF09,RasY09,Yua10,FatZ18},
as new analyses become available for the closely related
$\ell_1$-penalized LASSO regression in vector spaces.

Among these, our
method inherits the spirit of Reference \cite{MeiB06,Yua10} in
particular, where the authors considered the problem~\eqref{eq:glasso}
in terms of a collection of local regression problems defined for each
random variables.

\subsection*{LASSO and SLOPE}

Under a linear data model $b = A\beta + \epsilon$ with a data matrix $A \in \R^{n\times p}$ and $\epsilon \sim \mathcal N(0, \sigma^2 I_n)$, the~$\ell_1$-penalized estimation of $\beta$ is known as the LASSO regression~\cite{Tib96}, whose estimate is given by solving the following convex optimization problem:
\begin{equation*}
  \hat\beta = \argmin_{\beta \in \R^p} \;\; \frac12 \|b - A\beta\|_2^2 + \lambda \|\beta\|_1
\end{equation*}
where $\lambda >0$ is a tuning parameter which determines the sparsity of the estimate $\hat\beta$. Important statistical properties of $\hat\beta$ is that (i) the distance between the estimate $\hat\beta$ and the population parameter vector $\beta$ and (ii) the detection of non-zero locations of $\beta$ by $\hat\beta$. We can rephrase the former as the estimation error and the latter as the model selection error. These two types of error are dependent on how to choose the value of the tuning parameter $\lambda$. Regarding the model selection, it is known that the LASSO regression can control the family-wise error rate (FWER) at level $\alpha \in [0,1]$ by choosing $\lambda = \sigma \Phi^{-1} (1-\alpha /2p)$~\cite{BogB15}. The FWER is essentially the probability of including at least one entry as non-zero in $\hat\beta$ which is zero in~$\beta$. In high-dimensional cases where $p \gg n$, controlling FWER is quite restrictive since it is unavoidable to do false positive detection of nonzero entries. As a result, FWER control can lead to weak detection power of nonzero entries.

The SLOPE is an alternative procedure for the estimation of $\beta$, using the sorted $\ell_1$-norm penalization instead. The SLOPE solves a modified convex optimization problem, 
\begin{equation*}
   \hat\beta = \argmin_{\beta \in \R^p} \;\; \frac12 \|b-A\beta\|_2^2 + J_\lambda(\beta)
\end{equation*}
where $J_\lambda(\cdot)$ is the sorted $\ell_1$-norm defined by 
$$
J_\lambda(\beta) := \sum_{i=1}^p \lambda_i |\beta|_{(i)}
$$
where $\lambda_1>\lambda_2>\cdots>\lambda_p > 0$ and $|\beta|_{(k)}$ is the $k$th largest component of $\beta$ in magnitude. In Reference \cite{BogB15}, it has been shown that, for linear regression, the SLOPE procedure can control the false discovery rate (FDR) at level $q \in [0,1]$ of model selection by choosing $\lambda_i = \Phi^{-1}(1- i \cdot q/2p)$.
The FDR is the expected ratio of false discovery (i.e., the number of false nonzero entries) to total discovery. Since controlling FDR is less restrictive for model selection compared to the FWER control, FDR control can lead to a significant increase in detection power, while it may slightly increase the total number of false discovery~\cite{BogB15}.

\textls[-20]{This paper is motivated by the SLOPE method~\cite{BogB15} for its use of the SL1 regularization, where~it brings many benefits not available with the popular $\ell_1$-based regularization—the~capability of false discovery rate (FDR) control~\cite{BogB15,BrzS15}, adaptivity~to unknown signal sparsity~\cite{SuC16} and~clustering of coefficients~\cite{BonR08,FigN16}. Also, efficient optimization methods~\cite{BogB15,LeeB16,ZhaF18} and more theoretical analysis~\mbox{\cite{SuC16,CheB16,BelLT17,Der18}} are under active~research.}

In this paper, we propose a new procedure to find a sparse inverse
covariance matrix estimate, we~call \gslope (neighborhood selection Sorted L-One
Penalized Estimation). Our \gslope procedure uses the sorted
$\ell_1$-norm for penalized model selection, whereas the existing
\glasso~\eqref{eq:glasso} method uses the $\ell_1$-norm for the
purpose. We investigate our method in two aspects in theory and in
experiments, showing that (i) how the estimation error can be bounded,
and (ii) how the model selection (more specifically, the neighborhood
selection~\cite{MeiB06}) can be done with an FDR control in the edge
structure of the Gaussian Markov random field. We also provide an
efficient but straightforward estimation algorithm which fits for
parallel computation.

\section{\gslope (Neighborhood Selection Sorted L-One Penalized Estimation)}

Our method is based on the idea that the estimation of the inverse
covariance matrix of a multivariate normal distribution
$\mathcal N(0, \Sigma_{p\times p})$ can be decomposed into the
multiple regression on conditional
distributions~\cite{MeiB06,Yua10}. 

For a formal description of our method, let us consider a
$p$-dimensional random vector
$\rx \sim \mathcal N(0, \Sigma_{p\times p})$, denoting its $i$th
component as $\rx_i \in \R$ and the sub-vector without the $i$th
component as $\rx_{-i} \in \R^{p-1}$. For the inverse covariance
matrix $\Theta := \Sigma^{-1}$, we use $\Theta_{i}$ and $\Theta_{-i}$
to denote the $i$th column of the matrix and the rest of $\Theta$
without the $i$th column, respectively.

From the Bayes rule, we can decompose the full
log-likelihood into the following parts:
\begin{align*}
&\sum_{j=1}^n \log P_\Theta(\rx=x^j) = \sum_{j=1}^n \log P_{\Theta_i}(\rx_i = x_i^j | \rx_{-i} = x_{-i}^j) \\ 
&\qquad \qquad+ \sum_{j=1}^n \log P_{\Theta_{-i}}(\rx_{-i} = x_{-i}^j) .
\end{align*}

This decomposition allows us a block-wise optimization of the full
log-likelihood, which iteratively optimizes
$P_{\Theta_i}(\rx_i|\rx_{-i})$ while the parameters in
$P_{\Theta_{-i}}(\rx_{-i})$ are fixed at the current value.

\subsection{Sub-Problems}

In the block-wise optimization approach we mentioned above, we need to
deal with the conditional distribution
$P_{\Theta_i}(\rx_i | \rx_{-i})$ in each iteration. When
$\rx \sim \mathcal N(0, \Sigma)$, the conditional distribution also
follows the Gaussian distribution~\cite{And03}, in particular:
\begin{align*}
\rx_i | \rx_{-i} \sim \mathcal N(\mu_i , \sigma_i^2), \;
\begin{cases} \mu_i := \rx_{-i} \Sigma_{-i,-i}^{-1} \Sigma_{-i,i}\\
  \sigma_i^2 := \Sigma_{ii} - \Sigma_{i,-i}\Sigma_{-i,-i}^{-1}\Sigma_{-i,i} .
\end{cases}
\end{align*}

Here $\Sigma_{-i,i} \in \R^{p-1}$ denotes the $i$th column of $\Sigma$
without the $i$th row element,
$\Sigma_{-i,-i} \in \R^{(p-1)\times (p-1)}$ denotes the sub-matrix of
$\Sigma$ without the $i$th column and the $i$th row and
$\Sigma_{ii} = \Sigma_{i,i}$ is the $i$th diagonal component of
$\Sigma$. Once we define $\beta^i$ as follow,
$$
 \beta^i := \Sigma_{-i,-i}^{-1} \Sigma_{-i,i} \in \R^{p-1},
$$
then the conditional distribution now looks like:
\begin{align*}
\rx_i | \rx_{-i} \sim N(\mu_i, \sigma_i^2), \;\;
\begin{cases} \mu_i = \rx_{-i} \beta^i \\
  \sigma_i^2 = \Sigma_{ii} - (\beta^i)'\Sigma_{-i,-i} \beta^i .
\end{cases}
\end{align*}

This indicates that the minimization of the conditional log-likelihood
can be understood as a local regression for the random variable $\rx_i$,
under the data model:
\begin{equation}\label{eq:localreg}
 \rx_i = \rx_{-i} \beta^i + \nu_i, \;\; \nu_i \sim \mathcal N(0, \sigma_i^2) .
\end{equation}


To obtain the solution of the local regression
problem~\eqref{eq:localreg},
we consider a convex optimization based on the sorted
$\ell_1$-norm~\cite{BogB15,FigN16}.
In particular,  
for each local problem index $i \in \{1,\dots,p\}$, we solve
\begin{equation}\label{eq:slope2}
   \hat\beta^i \in \min_{\beta^i \in \R^{p-1}} \frac{1}{2} \|X_i - X_{-i}\beta^i\|_2^2 + \hat\sigma_i \cdot  J_\vlam (\beta^i) .
 \end{equation}
 
 Here, the matrix $X \in \R^{n\times p}$ consists of $n$
 i.i.d. $p$-dimensional samples in rows, $X_i$ is the $i$th column of
 $X$ and $X_{-i} \in \R^{n\times (p-1)}$ is the sub-matrix of $X$
 without the $i$th column. Note that the sub-problem~\eqref{eq:slope2}
 requires an estimate of $\sigma_i$, namely $\hat\sigma_i$, which will
 be computed dependent on the information on the other sub-problem
 indices (namely, $\Sigma_{-i,-i}$). Because of this, the
 problems~\eqref{eq:slope2} for indices $i=1,2,\dots,p$ are not
 independent. This contrasts our method to the neighborhood
 selection algorithms~\cite{MeiB06,Yua10}, based on the
 $\ell_1$-regularization.

\subsection{Connection to  Inverse Covariance Matrix Estimation}

With $\hat\beta^i$ obtained from \eqref{eq:slope2} as an estimate of
$\Sigma^{-1}_{-i,-i}\Sigma_{-i,i}$ for $i=1,2,\dots,p$, now the
question is how to use the information for the estimation of
$\Theta=\Sigma^{-1}$ without an explicit inversion of the matrix.  For~the purpose, we first consider the block-wise formulation of
$\Sigma \Theta = I$:
$$
\begin{bmatrix}
  \Sigma_{-i,-i} & \Sigma_{-i,i} \\
  \Sigma_{i, -i} & \Sigma_{ii}
\end{bmatrix}
\begin{bmatrix}
  \Theta_{-i,-i} & \Theta_{-i,i} \\
  \Theta_{i,-i} & \Theta_{ii}
\end{bmatrix}
=
\begin{bmatrix}
  I_{(p-1)\times (p-1)} & 0 \\
    0 & 1
\end{bmatrix}
$$
putting the $i$th row and the $i$th column to the last positions
whenever necessary. There~we can see that
$\Sigma_{-i,-i}\Theta_{-i,i} + \Sigma_{-i,i}\Theta_{ii} = 0$. Also,
from the block-wise inversion of $\Theta = \Sigma^{-1}$, we have
(unnecessary block matrices are replaced with $*$):
$$
\begin{aligned}
& \begin{bmatrix}
  \Sigma_{-i,-i} & \Sigma_{-i,i} \\
  \Sigma_{i, -i} & \Sigma_{ii}
\end{bmatrix}^{-1}
&=
\begin{bmatrix}
  ~~* 
  & * 
  \\
  ~~* 
  & ( \Sigma_{ii} - \Sigma_{i, -i} \Sigma_{-i,-i}^{-1} \Sigma_{-i,i} )^{-1}
\end{bmatrix}
\end{aligned}
$$

From these and the definition of
$\beta^i$, we can establish the following relations:
\begin{equation}\label{eq:sub}
\begin{aligned}
\begin{cases}
  \beta^i &= \Sigma_{-i,-i}^{-1} \Sigma_{-i,i} = - \frac{\Theta_{-i,i}}{\Theta_{ii}} \\
  \Theta_{ii} &= (\Sigma_{ii} - (\beta^i)' \Sigma_{-i,-i}\beta^i )^{-1} = 1/\sigma_i^2 .
\end{cases}
\end{aligned}
\end{equation}

Note that $\Theta_{ii}>0$ for a positive definite matrix
$\Theta$ and that \eqref{eq:sub} implies the sparsity pattern
of $\Theta_{-i,i}$ and $\beta^i$ must be the same.
Also, the updates~\eqref{eq:sub}
for $i=1,2,\dots,i$ are not independent, since~the computation of
$\sigma_i = 1/\Theta_{ii}$ depends on
$\Sigma_{-i,-i}$. However, if we estimate
$\sigma_i$ based on the sample covariance matrix instead of
$\Sigma$, (i) our updates~\eqref{eq:sub} no longer needs to explicitly
compute or store the $\Sigma_{p\times
  p}$ matrix, unlike~the \glasso and (ii) our sub-problems become
mutually independent and thus solvable in~parallel.

\section{Algorithm}
Our algorithm, called \gslope, is summarized in
Algorithm~\ref{alg:gslope}, which is essentially a block-coordinate descent algorithm~\cite{BecT13,TreT14}. Our algorithm may look similar to that of
\citet{Yua10} but there are several important differences—(i) each
sub-problem in our algorithm solves a SLOPE formulation (SL1
regularization), while Yuan's sub-problem is either LASSO or Dantzig
selector ($\ell_1$~regularization); (ii) our sub-problem makes use of
the estimate $\hat\Theta_{ii}$ in addition to
$\hat\Theta_{-i,i}$.

\vspace{12pt}
\begin{algorithm}[H]
\caption{The \gslope Algorithm}\label{alg:gslope}
  \KwIn{$X \in \R^{n\times p}$ with zero-centered columns}
  \KwIn{$S = \frac1n X' X$}
  \KwIn{The target level of FDR $q$} 
  Set $\lambda=(\lambda_1,\dots,\lambda_{p-1})$ according to Section~\ref{sec:lambda}\;
\While{$\hat\Theta$ not converged}{
 \For{$i=1,2,\dots,p$}{
      Compute $\hat\beta^i$ as the solution to the sub-problem~\eqref{eq:slope2}:
      $$
       \hat\beta^i = \argmin_{\beta \in \R^{p-1}} \; \frac12 \|X_i - X_{-i}\beta\|^2 + (\hat\Theta_{ii})^{-1/2} J_\lambda(\beta)
      $$
      Update $\hat\Theta_{ii}$ and $\hat\Theta_{-i,i}$  (\eqref{eq:sub} and Section~\ref{sec:thetaii}):
       \begin{align*}\begin{cases}
         \hat\Theta_{ii} &= (S_{ii} - 2(\hat\beta^i)' S_{-i,i} + (\hat\beta^i)' S_{-i,-i}\hat\beta^i)^{-1}\\
         \hat\Theta_{-i,i} &= -\hat\Theta_{ii}\hat\beta^i
         \end{cases}
       \end{align*}
      }

      Make $\hat\Theta$ a symmetric matrix:
       $$
          \hat\Theta \leftarrow \argmin_{\Theta=\Theta'} \;\; \|\Theta - \hat\Theta\|_F^2
          $$
}
\end{algorithm}

\subsection{Sub-Problem Solver}
To solve our sub-problems~\eqref{eq:slope2}, we use the \textrm{SLOPE}
R package, which implements the proximal gradient descent
algorithm of Reference \cite{BecT09} with acceleration based on Nesterov's
original idea~\cite{Nes83}. The~algorithm requires to compute the
proximal operator involving $J_\lambda(\cdot)$, namely
$$
 \prox_{J_\lambda} (z) := \argmin_{x \in \R^{p-1}} \;\; \frac12 \|x-z\|^2 + J_\lambda(x) .
$$

This can be computed in $\cO(p\log p)$ time using an algorithm from Reference
\cite{BogB15}. The optimality of sub-problem is declared by the
primal-dual gap and we use a tight threshold value, $10^{-7}$.

\subsection{Choice of $\lambda = (\lambda_1, \dots, \lambda_{p-1})$}\label{sec:lambda}
For the sequence of $\lambda$ values in sub-problems, we use so-called
the Benjamini-Hochberg (BH) sequence~\cite{BogB15}:
\begin{equation}\label{eq:bh}
\lambda_i^{BH} = \Phi^{-1} (1- iq / 2(p-1)), \; i=1,\dots,p-1 .
\end{equation}

Here $q \in [0,1]$ is the target level of FDR control we discuss
later and $\Phi^{-1}(\alpha)$ is the $\alpha$th quantile of the
standard normal distribution. In fact, when the design matrix $X_{-i}$
in the SLOPE sub-problem~\eqref{eq:slope2} is not orthogonal, it is
beneficial to use an adjusted version of this sequence. This sequence
is generated~by:
$$
\lambda_i = \lambda_i^{BH} \sqrt{1 + w(i-1) \sum_{j<i} \lambda_j^2}, \;\; w(k) = \frac{1}{n-k-1},
$$
and if the sequence $\lambda_1, \dots, \lambda_k$ is non-increasing
but $\lambda_i$ begins to increase after the index $k^*$, we set all
the remaining values equal to $\lambda_{k^*}$, so that the resulting
sequence will be non-increasing. For more discussion about the
adjustment, we refer to Section 3.2.2 of Reference \cite{BogB15}.

\subsection{Estimation of $\Theta_{ii}$}\label{sec:thetaii}
To solve the $i$th sub-problem in our algorithm, we need to estimate the
value of $\Theta_{ii}$. This can be done using
\eqref{eq:sub}, that is,
$\Theta_{ii} = (\Sigma_{ii} - (\beta^i)' \Sigma_{-i,-i}\beta^i
)^{-1}$. However, this implies that (i) we need to keep an estimate of
$\Sigma \in \R^{p\times p}$ additionally, (ii) the computation of the
$i$th sub-problem will be dependent on all the other indices, as it
needs to access $\Sigma_{-i,-i}$, requiring the algorithm to run
sequentially.

To avoid these overheads, we compute the estimate using the sample
covariance matrix $S = \frac1n X'X$ instead (we assume that the columns of
$X\in\R^{n\times p}$ are centered):
$$
\tilde \Theta_{ii} = (S_{ii} - 2\hat\beta' S_{-i,i} + \hat\beta' S_{-i,-i}\hat\beta)^{-1} .
$$

This allows us to compute the inner loop of
Algorithm~\ref{alg:gslope} in parallel.


\subsection{Stopping Criterion of the External Loop}
To terminate the outer loop in Algorithm~\ref{alg:gslope}, we check if
the diagonal entries of $\hat\Theta$ have converged, that is, the
algorithm is stopped when the $\ell_\infty$-norm different between two
consecutive iterates is below a threshold value, $10^{-3}$. The value
is slightly loose but we have found no practical difference by making
it tighter. Note that it is suffice to check the optimality of the
diagonal entries of $\hat\Theta$ since the optimality of
$\hat\beta^i$'s is enforced by the sub-problem solver and
$\hat\Theta_{-i,i} = -\hat\Theta_{ii}\hat\beta^i$.


\subsection{Uniqueness of Sub-Problem Solutions}
When $p>n$, our sub-problems may have multiple solutions, which may
prevent the global convergence of our algorithm. We may adopt the
technique in Reference \cite{RazHL13} to inject a strongly convex proximity term
into each sub-problem objective, so that each sub-problem will have a
unique solution. In our experience, however, we encountered no
convergence issues using stopping threshold values in the range of
$10^{-3} \sim 10^{-7}$ for the outer loop.

\section{Analysis}

In this section, we provide two theoretical results of our \gslope
procedure—(i) an estimation error bound regarding the distance
between our estimate $\tilde\Theta$ and the true model parameter
$\Theta$ and~(ii)~group-wise FDR control on discovering the true edges
in the Gaussian MRF corresponding to~$\Theta$.

We first discuss about the estimation error bound, for which we
divide our analysis into two parts regarding (i) off-diagonal entries
and (ii) diagonal entries of~$\Theta$.

\subsection{Estimation Error Analysis}
\unskip
\subsubsection{Off-Diagonal Entries}


From \eqref{eq:sub}, we see that
$\hat\beta^i = -\hat\Theta_{-i,i} / \hat\Theta_{ii}$, in other words,
when $\hat\Theta_{ii}$ is fixed, the off-diagonal entries
$\hat\Theta_{-i,i}$ is determined solely by $\hat\beta^i$, and therefore we
can focus on the estimation error of $\hat\beta^i$.

To discuss the estimation error of $\hat\beta^i$, it is convenient to
consider a constrained reformulation of the sub-problem~\eqref{eq:slope2}:
\begin{equation}\label{eq:slope.const}
\begin{aligned}
  \hat\beta \in \argmin_{\beta \in \R^d} \;\; J_\vlam(\beta) \;\;
  \text{s.t.} \;\; \frac{1}{n} \|b  - A \beta \|_1 \le \epsilon .
\end{aligned}
\end{equation}

Hereafter, for the sake of simplicity, we use notations $b := X_i \in \R^n$
and $A := X_{-i} \in R^{n\times d}$ for $d:=p-1$, also dropping
sub-problem indices in $\hat\beta$ and $\epsilon>0$.
In this view, the data model being considered in each sub-problem is as
follows,
$$
  b = A\beta^* + \nu, \;\; \nu \sim \mathcal N(0, \sigma_i^2  I_n) .
$$
  
For the analysis, we make the following assumptions:
\begin{enumerate}

\item The true signal $\beta^* \in \R^d$ satisfies
  $\|\beta^*\|_1 \le \sqrt{s}$ for some $s>0$ (this condition is
  satisfied for example, if~$\|\beta^*\|_2 \le 1$ and $\beta^*$ is $s$-sparse,
  that is, it has at most $s$ nonzero elements).

  \item   The noise satisfies the condition
  $ \frac{1}{n} \|\nu\|_1 \le \epsilon.  $ This will allow us to say
  that the true signal $\beta^*$ is feasible with respect to the
  constraint in \eqref{eq:slope.const}.
\end{enumerate}
 
 We provide the following result, which shows
that $\hat\beta$ approaches the true $\beta^*$ in high probability:
%


\medskip
\begin{Theorem}\label{thm:pb.slope} Suppose that $\hat\beta$ is
  an estimate of $\beta^*$ obtained by solving the
  sub-problem~\eqref{eq:slope.const}. Consider the factorization
  $A := X_{-i} = BC$ where $B\in\R^{n\times k}$ ($n \le k$) is a
  Gaussian random matrix whose entries are sampled i.i.d. from
  $\mathcal N(0,1)$ and $C \in \R^{k\times d}$ is a deterministic
  matrix such that $C' C = \Sigma_{-i,-i}$. Such decomposition is
  possible since the rows of $A$ are independent samples from
  $\mathcal N(0, \Sigma_{-i,-i})$. Then we have,
  \begin{align*}
    \| \hat\beta - \beta^* \|_C^2
    \le \sqrt{2\pi} \left( 8\sqrt{2}\|C\|_1\frac{\lambda_1}{\bar\lambda} \sqrt{\frac{s\log k}{n}}  + \epsilon + t \right)
  \end{align*}
  with probability at least $1-2\exp\left( -\frac{nt^2}{2s} \frac{\bar\lambda^2}{\lambda_1^2}
  \right)$, where  $\bar\lambda = \frac{1}{d}\sum_{i=1}^d \lambda_i$
  and   $\|C\|_1 = \max_{j=1,\dots,d} \sum_{i=1}^k |C_{ij}|$.
\end{Theorem}
\bigskip

   We need to discuss a few results before proving Theorem~\ref{thm:pb.slope}.


\medskip
\begin{Theorem}\label{thm:pb.u}
  Let $T$ be a bounded subset of $\R^d$. For an $\epsilon>0$, consider the set
  $$
     T_\epsilon := \left\{ u \in T: \frac{1}{n} \|A u\|_1 \le \epsilon \right\} .
  $$
  
  Then
  $$
     \sup_{u\in T_{\epsilon}} (u' C' C u)^{1/2} \le \sqrt{\frac{8\pi}{n}} \E\sup_{u\in S} |\langle C' g, u\rangle | + \sqrt{\frac{\pi}{2}} (\epsilon + t),
  $$
  holds with probability at least
  $1-2\exp\left( -\frac{nt^2}{2 d(T)^2} \right)$, where $g \in \R^k$
    is a standard Gaussian random vector and $d(T) := \max_{u\in T} \|u\|_2$.
 \end{Theorem}

\begin{proof}
  The result follows from an extended general $M^*$ inequality in
  expectation~\cite{FigN14}.
\end{proof}

The next result shows that the first term of the upper bound in
Theorem~\ref{thm:pb.u} can be bounded without using expectation.
\begin{Lemma}\label{thm:width}
 The quantity $\E \sup_{u \in \cK - \cK} | \langle C' g, u \rangle |$
 is called the {\em width} of $\cK$ and is bounded as follows,
$$
\E \sup_{u \in \cK - \cK} |\langle C' g, u \rangle | \le 4\sqrt{2} \|C\|_1 \frac{\lambda_1}{\bar\lambda} \sqrt{s \log k} .
$$
\end{Lemma}
\begin{proof}
 This result is a part of the proof for Theorem 3.1 in Reference \cite{FigN14}.
\end{proof}

Using Theorem~\ref{thm:pb.u} and Lemma~\ref{thm:width}, we can derive
a high probability error bound on the estimation from noisy
observations,
$$
  b = A\beta^* + \nu, \;\; \frac{1}{n} \|\nu\|_1 \le \epsilon,
$$
\textls[10]{where the true signal $\beta^*$ belongs to a bounded subset
$\cK \in \R^d$. The following corollaries are straightforward
extensions of Theorems 3.3 and 3.4 of Reference \cite{FigN14}, given our
Theorem~\ref{thm:pb.u} (so we skip the~proofs).}

\begin{Corollary}\label{thm:pb.est}
  Choose $\hat\beta$ to be any vector satisfying that
  $$
    \hat\beta \in \cK, \;\; \text{and} \;\; \frac{1}{n} \|b - A\hat\beta \|_1 \le \epsilon.
  $$
  
  Then,
  \begin{align*}
  &\sup_{\beta \in \cK} [ (\hat\beta - \beta)' C' C (\hat\beta - \beta)]^{1/2}\\
  & \;\; \le \sqrt{2\pi} \left( 8\sqrt{2}\|C\|_1\frac{\lambda_1}{\bar\lambda} \sqrt{\frac{s\log k}{n}}  + \epsilon + t \right)
  \end{align*}
  with probability at least $1-2\exp\left( -\frac{2nt^2}{d(T)^2} \right)$.
\end{Corollary}

Now we show the error bound for the estimates we obtain by solving the
optimization problem~\eqref{eq:slope.const}. For the purpose, we make
use of the Minkowski functional of the set $\cK$,
$$
 \|\beta\|_\cK := \inf\{r > 0 : r \beta \in \cK\}
$$
If $\cK \subset \R^p$ is a compact and origin-symmetric convex set
with non-empty interior, then $\|\cdot\|_\cK$ defines a norm in
$\R^p$. Note that $\beta \in \cK$ if and only if
$\|\beta\|_\cK \le 1$.

\begin{Corollary}\label{thm:pb.opt}
  $$
  \hat\beta \in \argmin_{\beta \in \R^p} \; \|\beta\|_\cK \;\; \text{subject to} \;\; \frac{1}{n} \|b - A\beta\|_1 \le \epsilon.
  $$
  
  Then
  \begin{align*}
    &\sup_{\beta \in \cK} [ (\hat\beta - \beta)' C' C (\hat\beta - \beta)]^{1/2}\\
    &\;\; \le \sqrt{2\pi} \left( 8\sqrt{2}\|C\|_1\frac{\lambda_1}{\bar\lambda} \sqrt{\frac{s\log k}{n}}  + \epsilon + t \right)
  \end{align*}
    with probability at least $1-2\exp\left( -\frac{2nt^2}{d(T)^2} \right)$.
\end{Corollary}

Finally, we show that solving the constrained form of the
sub-problems~\eqref{eq:slope.const} also satisfies essentially the same
error bound in Corollary~\ref{thm:pb.opt}.

\begin{proof}[Proof of Theorem~\ref{thm:pb.slope}]
  Since we assumed that $\|\beta\|_1 \le \sqrt{s}$, we construct the
  subset $\cK$ so that all vectors $\beta$ with $\|\beta\|_1\le \sqrt{s}$
  will be contained in $\cK$. That is,
$$
 \cK := \{ \beta \in \R^p : J_\lambda (\beta) \le \lambda_1 \sqrt{s} \} .
$$

This is a sphere defined in the SL1-norm $J_\lambda(\cdot)$: in this
case, the Minkowski functional $\|\cdot\|_\cK$ is proportional to
$J_\lambda(\cdot)$ and thus the same solution minimizes
both $J_\lambda(\cdot)$ and $\|\cdot\|_\cK$.

Recall that $d(T) := \max_{u\in T} \|u\|_2$ and we choose
$T = \cK - \cK$. Since
$\|\beta\|_2 \le \|\beta\|_1 \le \frac{1}{\bar\lambda}
J_\lambda(\beta)$, for~$\beta \in \cK$ we have
$\|\beta\|_2 \le \frac{\lambda_1}{\bar\lambda} \sqrt{s}$. This implies that
$d(T) = 2\frac{\lambda_1}{\bar\lambda} \sqrt{s}$.

\end{proof}

\subsubsection{Diagonal Entries}

We estimate $\Theta_{ii}$ based on the residual sum of squares (RSS)
as suggested by \cite{Yua10},
\begin{align*}
 \hat\sigma_i^2 = (\hat\Theta_{ii})^{-1} &= \| X_i - X_{-i}' \hat\beta^i \|^2 / n \\
&= S_{ii}
- 2 (\hat\beta^i)' S_{-i,i} + (\hat\beta^i)' S_{-i,-i} \hat\beta^i .
\end{align*}

Unlike in Reference \cite{Yua10}, we directly analyze the estimation error of
$\hat\Theta_{ii}$ based on a chi-square tail bound.
\begin{Theorem}\label{thm:pb.diag}
  For all small enough $\alpha > 0$ so that
  $\alpha/\sigma_i^2 \in [0,1/2)$, we have
  $$
  \Prob( |  \hat\Theta_{ii}^{-1}  - \Theta_{ii}^{-1} -  \delta((\beta^i)^*, \hat\beta^i)  | \ge \alpha ) \le \exp\left(-\frac{3}{16} n \alpha^2 \right)
  $$
  where for $\nu \sim \cN(0, \sigma_i^2 I_n)$,
  $$
  \delta(\beta^*, \hat\beta) :=  (\beta^* - \hat\beta)' S_{-i,-i} (\beta^* - \hat\beta) + 2 \nu' X_{-i}(\beta^* -\hat\beta) / n 
  $$
\end{Theorem}
\begin{proof}
  Using the same notation as the previous section, that is, $b = X_i$ and
  $A = X_{-i}$, consider the estimate in discussion,
  $$
    \hat\sigma_i^2 = \|b - A\hat\beta\|^2 / n = \|A(\beta^*-\hat\beta) + \nu\|^2 / n
  $$
  where the last equality is from $b = A\beta^* + \nu$.
  Therefore,
  $$
   \hat\sigma_i^2  = (\beta^* - \hat\beta)' S_{-i,-i} (\beta^* - \hat\beta) + 2 \nu' A(\beta^* -\hat\beta) / n +  \nu'\nu / n
  $$
  where $S_{-i,-i} := A'A/n = X_{-i}'X_{-i}/n$. The last term is the
  sum of squares of independent $\nu_i \sim \cN(0, \sigma_i^2)$ and 
  therefore it follows the chi-square distribution, that is,
  $(\nu'\nu)/\sigma_i^2 \sim  \chi^2_n$.  Applying the tail bound~\cite{Joh01}
  for a chi-square random variable $Z$ with $d$ degrees of freedom:
  $$
  \Prob( |d^{-1} Z - 1| \ge \alpha ) \le \exp\left(-\frac{3}{16} d \alpha^2 \right), \;\; \alpha \in [0, 1/2) .
  $$
  we get for all small enough $\alpha>0$,
  $$
  \Prob( |\nu'\nu / n - \sigma_i^2| \ge \alpha ) \le \exp\left(-\frac{3}{16} n \alpha^2 \right) .
  $$
\end{proof}

\subsubsection{Discussion on Asymptotic Behaviors}\label{sec:est.dis}

Our two main results above, Theorems~\ref{thm:pb.slope} and
\ref{thm:pb.diag}, indicate how well our estimate of the off-diagonal
entries $\hat\Theta_{-i,i}$ and diagonal entries $\hat\Theta_{ii}$
would behave. Based on these results, we can discuss the estimation
error of the full matrix $\hat\Theta$ compared to the true precision
matrix $\Theta$.


From Theorem~\ref{thm:pb.slope}, we can deduce that with $C'C = \Sigma_{-i,-i}$ and $z(n) = O(\sqrt{s\log k/n})$,
\begin{align*}
  \text{ev}_{\min}(\Sigma_{-i,-i}) \|\hat\beta - \beta^*\|
  &\le [(\hat\beta - \beta^*)\Sigma_{-i,-i} (\hat\beta-\beta)]^{1/2} \\
  &\le \sqrt{2\pi}(z(n)+\epsilon+t)
\end{align*}
where $\text{ev}_{\min}(\Sigma_{-i,-i})$ is the smallest eigenvalue of
the symmetric positive definite $\Sigma_{-i,-i}$. That~is, using~the
interlacing property of eigenvalues, we have
   $$
    \|\hat\beta - \beta^*\| \le \frac{\sqrt{2\pi} (z(n)+\epsilon+t)}{\text{ev}_{\min}(\Sigma_{-i,-i})} \le \sqrt{2\pi}\|\Theta\| (z(n)+\epsilon+t)
   $$
   where $\|\Theta\| = \text{ev}_{\max}(\Theta)$ is the spectral
   radius of $\Theta$. Therefore when $n\to\infty$, the distance
   between $\hat\beta$ and $\beta^*$ is bounded by
   $\sqrt{2\pi}\|\Theta\|(\epsilon+t)$. Here, we can consider
   $t \to 0$ in a way such as $n t^2 \to \infty$ as $n$ increases, so
   that the bound in Theorem~\ref{thm:pb.slope} will hold with the
   probability approaching one. That is, in~rough asymptotics,
   \begin{equation}\label{eq:asym.1}
   \|\hat\beta - \beta^*\| \le \epsilon \sqrt{2\pi} \|\Theta\| .
   \end{equation}
   
   Under the conditions above, Theorem~\ref{thm:pb.diag} indicates that:
   \begin{align*}
     \delta(\beta^*, \hat\beta) &\le \|S\| \|\beta^* - \hat\beta\|^2 + 2 \|\nu/n\|_1 \|X_{-i} (\beta^* - \hat\beta)\|_\infty \\
        & \le  2\epsilon^2 \|\Theta\| (\pi \|S\| \|\Theta\| + \sqrt{2\pi} \|X_{-i}\|_\infty) 
   \end{align*}
   using our assumption that $\|\nu/n\|_1 \le \epsilon$. We can
   further introduce assumptions on $\|\Theta\|$ and $S\Theta$ as in Reference~\cite{Yua10}, so that we can quantify the upper bound but here we
   will simply say that $\delta(\beta^*,\hat\beta) \le 2c\epsilon^2$,
   where $c$ is a constant depending on the properties of the full
   matrices $S$ and $\Theta$. If this is the case, then~from
   Theorem~\ref{thm:pb.diag} for an $\alpha \to 0$ such that
   $n\alpha^2 \to \infty$, we see that
   \begin{equation}\label{eq:asym.2}
      | \hat\Theta_{ii}^{-1} - \Theta_{ii}^{-1} | \le \delta((\beta^i)^*, \hat\beta^i) \le 2c\epsilon^2 ,
   \end{equation}
    with the probability approaching one.

    Therefore, if we can drive $\epsilon \to 0$ as $n\to\infty$, then
    \eqref{eq:asym.1} and \eqref{eq:asym.2} imply the convergence of
    the diagonal $\hat\Theta_{ii} \to \Theta_{ii}$ and the
    off-diagonal $\hat\Theta_{-i,i} \to \Theta_{-i,i}$ in
    probability. Since $\|X_i - X_{-i}\beta\|_1 \le \epsilon$ (or
    $\|X_i - X_{-i}\beta^i\| \le \epsilon_i$ more precisely),
    $\epsilon \to 0$ is likely to happen when $(p-1)/n \ge 1$,
    implying that $p \to \infty$ as well as $n \to \infty$.

\subsection{Neighborhood FDR Control under Group Assumptions}

%

Here we consider $\hat\beta$ obtained by solving the unconstrained
form~\eqref{eq:slope2} with the same data model we discussed above for
the sub-problems:
\begin{equation}\label{eq:reg.sub}
  b = A\beta + \nu, \;\; \nu \sim \cN(0, \sigma^2 I_n)
\end{equation}
with $b=X_i \in \R^n$ and $A=X_{-i} \in \R^{n\times d}$
with $d=p-1$. Here we focus on a particular but an interesting case where
the columns of $A$ form orthogonal groups, that is, under the
decomposition $A = BC$, $C'C = \Sigma_{-i,-i}$ forms a block-diagonal
matrix. We also assume that the columns of $A$ belonging to the same
group are highly correlated, in the sense that for any columns $a_i$
and $a_j$ of $A$ corresponding to the same group, their correlation is
high enough to satisfy that
$$
 \|a_i - a_j \| \le \min_{i=1,\dots,d-1}\{\lambda_i - \lambda_{i+1}\} / \|b\| .
$$

This implies that $\hat\beta_i = \hat\beta_j$ by Theorem 2.1 of Reference
\cite{FigN16}, which simplifies our analysis. Note that if $a_i$ and
$a_j$ belong to different blocks, then our assumption above implies
$a_i'a_j = 0$. Finally, we further assume that $\|a_i\| \le 1$ for
$i=1,\dots,d$.

Consider a collection of non-overlapping index subsets
$g \subset \{1,\dots,d\}$ as the set of groups $G$. Under the
block-diagonal covariance matrix assumption above, we see that
$$
  A\beta = \sum_{g \in G} \sum_{i\in g} a_i \beta_i = \sum_{g \in G} \left( \sum_{i\in g} a_i \right)  \beta_g = \sum_{g\in G} \frac{\tilde a_g}{\|\tilde a_g\|} \beta'_g
$$
where $\beta_g$ denotes the representative of the same coefficients
$\beta_i$ for all $i\in g$, $\tilde a_g := \sum_{i\in g} a_i$ and~$\beta'_g := \|\tilde a_g\| \beta_g$. This tells us that we can
replace $A\beta$ by $\tilde A \beta'$, if we define
$\tilde A \in \R^{n\times |G|}$ containing
$\frac{\tilde a_g}{\|\tilde a_g\|}$ in its columns (so that
$\tilde A' \tilde A = I_{|G|})$ and consider the vector of
group-representative coefficients $\beta' \in \R^{|G|}$.

The regularizer can be rewritten similarly,
$$
  J_\lambda(\beta) = \sum_{i=1}^{|G|} \lambda'_i |\beta'|_{(i)} = J_{\lambda'}(\beta')
$$
where
$ \lambda'_i := (\lambda_{\sum_{j=1}^{i-1}|g_{(j)}| + 1} + \cdots +
\lambda_{\sum_{j=1}^i |g_{(i)}|} ) / \|\tilde a_{g_{(i)}} \|
$,
denoting by $g_{(i)}$ the group which has the $i$th largest
coefficient in $\beta'$ in magnitude and by $|g_{(i)}|$ the size of
the group.

Using the fact that $\tilde A'\tilde A = I_{|G|}$, we can recast the
regression model \eqref{eq:reg.sub} as
$
 b' = \tilde A' b = \beta + \tilde A' \nu \; \sim \; \cN(\beta, \sigma^2 I_{|G|})
$,
and consider a much simpler form of the problem~\eqref{eq:slope2},
\begin{equation}\label{eq:slope.grp}
 \hat\beta' = \argmin_{\beta' \in \R^{|G|}} \;\; \frac12 \| b' - \beta' \|^2 + \sigma J_{\lambda'} (\beta') ,
\end{equation}
where we can easily check that
$\lambda' = (\lambda'_1, \dots, \lambda'_{|G|})$ satisfies
$\lambda'_1 \ge \dots \ge \lambda'_{|G|}$. This is exactly the form of
the SLOPE problem with an orthogonal design matrix in Reference \cite{BogB15},
except that the new $\lambda'$ sequence is not exactly the
Benjamini-Hochberg sequence~\eqref{eq:bh}.

We can consider the problem~\eqref{eq:slope2} (and respectively
\eqref{eq:slope.grp}) in the context of multiple hypothesis testing of
$d$ (resp. $|G|$) null hypothesis $H_i : \beta_i = 0$ (resp.
$H_{g_i}: \beta'_i = 0$) for $i=1,\dots,d$ (resp.~$i=1,\dots,|G|$),
where we reject $H_i$ (resp. $H_{g_i}$) if and only if $\hat\beta_i \neq 0$
(resp. $\hat\beta'_i \neq 0$). In~this setting, the~Lemmas B.1 and B.2 in Reference
\cite{BogB15} still holds for our problem~\eqref{eq:slope.grp}, since
they are independent of choosing the particular $\lambda$ sequence.

In the following, $V$ is the number of individual false rejections,
$R$ is the number of total individual rejections and $R_g$ is the
number of total group rejections. The following lemmas are slightly
modified versions of Lemmas B.1 and B.2 from Reference \cite{BogB15},
respectively, to fit our group-wise setup.

\begin{Lemma}\label{thm:fdr.lem1}
  Let $H_g$ be a null hypothesis and let $r \ge 1$. Then
  \begin{align*}
    &\{ b': H_g \; \text{is rejected} \, \wedge \, R_g=r \}
    = \{ b': |b'_g| > \sigma \lambda'_r \, \wedge \, R_g=r\} .
  \end{align*}
\end{Lemma}

\begin{Lemma}\label{thm:fdr.lem2}
  Consider applying the procedure~\eqref{eq:slope.grp} for new data
  $\tilde b' = (b'_1, \dots, b'_{g-1}, b'_{g+1}, \dots, b'_{|G|})$ with
  $\tilde \lambda = (\lambda'_2, \dots, \lambda'_d)$ and let
  $\tilde R_g$ be the number of group rejections made. Then with
  $r\ge 1$,
  \begin{align*}
    &\{ b': |b'_g| > \sigma \lambda'_r \, \wedge \, R_g=r \} 
     \subset \{ b': |b'_g| > \sigma\lambda'_r \, \wedge \, \tilde R_g = r-1 \}.
  \end{align*}
\end{Lemma}

Using these, we can show our FDR control result.
\begin{Theorem}\label{thm:fdr}
  Consider the procedure~\eqref{eq:slope.grp} we consider in a
  sub-problem of \gslope as a multiple testing of group-wise hypotheses,
  where we reject the null group hypothesis $H_{g}: \beta'_g = 0$ when
  $\beta'_g \neq 0$, rejecting~all individual hypotheses in the group,
  that is, all $H_i$, $i\in g$, are rejected. Using
  $\lambda_1,\dots,\lambda_{p-1}$ defined as in \eqref{eq:bh}, the~procedure controls FDR at the level $q\in [0,1]$:
$$
\text{FDR} = \E \left[ \frac{V}{R \vee 1}\right] \le \frac{G_0q}{d} \le q
$$
where
$$
\begin{cases}
  G_0 &:= |\{g: (\beta'_g)^* = 0 \}| \;\; \text{(\# true null group hypotheses)} \\
  V &:= |\{i: \beta^*_i = 0, \hat\beta^*_i \neq 0\}| \;\; \text{(\# false ind. rejections)}\\
  R &:= |\{i: \hat\beta^*_i \neq 0 \}| \;\; \text{(\# all individual rejections)}
\end{cases}
$$
\end{Theorem}
\begin{proof}
Suppose that $H_g$ is rejected. Then,
\begin{align*}
  &\Prob (H_g \; \text{rejected} \; \wedge \; R_g=r)
   \stackrel{(i)}{\le} \Prob (|b'_g| \ge \sigma\lambda'_r \; \wedge \; \tilde R_g = r-1) \\
  &\;\; \stackrel{(ii)}{=}  \Prob (|b'_g| \ge \sigma\lambda'_r) \; \Prob (\tilde R_g = r-1) \\
  &\;\; \stackrel{(iii)}{\le} \Prob \left(|b'_g| \ge \sigma \frac{|g_{(r)}| \lambda_{\sum_{j=1}^r |g_{(j)}|}}{\|\tilde a_{g_{(r)}}\|} \right) \; \Prob (\tilde R_g = r-1) \\
  &\;\; \stackrel{(iv)}{\le}  \Prob (|b'_g| \ge \sigma \lambda_{\sum_{j=1}^r |g_{(j)}|}) \; \Prob (\tilde R_g = r-1)
\end{align*}

The derivations above are—(i) by using Lemmas \ref{thm:fdr.lem1} and
\ref{thm:fdr.lem2}; (ii) from the independence between $b'_g$ and
$\tilde b'$; (iii) by taking the smallest term in the summation of
$\lambda'_r$ multiplied by the number of terms; (iv)~due to the
assumption that $\|a_i\|\le 1$ for all $i$, so that
$\|\tilde a_{g_{(r)}}\| \le |g_{(r)}|$ by triangle inequality.

Now, consider the group-wise hypotheses testing in \eqref{eq:slope.grp}
is configured in a way that the first $G_0$ hypotheses are null in
truth, that is, $H_{g_i}: \beta'_i = 0$ for $i\le G_0$.
Then we have:
\begin{align*}
  \text{FDR} &= \E \left[ \frac{V}{R \vee 1}\right]
             = \sum_{r=1}^{|G|} \E \left[ \frac{V}{\sum_{j=1}^r|g_{(j)}|} \vone_{\{R_g=r\}} \right]\\
             &=\sum_{r=1}^{|G|} \frac{1}{\sum_{j=1}^r|g_{(j)}|} \sum_{i=1}^{G_0} \E[ \vone_{\{H_{g_i} \text{ rejected}\}} \vone_{\{R_{g_i}=r\}} ]\\
             &=\sum_{r=1}^{|G|} \frac{1}{\sum_{j=1}^r|g_{(j)}|} \sum_{i=1}^{G_0} \Prob(H_{g_i} \text{ rejected} \wedge R_{g_i}=r ) \\
             &\le \sum_{r=1}^{|G|} \frac{1}{\sum_{j=1}^r|g_{(j)}|} \sum_{i=1}^{G_0}\frac{\sum_{j=1}^r |g_{(j)}| q}{d} \Prob (\tilde R_{g_i} = r-1)\\
  &= \sum_{r\ge 1} q \frac{G_0}{d} \Prob (\tilde R_{g_i} = r-1) =  q \frac{G_0}{d} \le q.
\end{align*}
\end{proof}

Since the above theorem applies for each sub-problem, which can be
considered as for the $i$th random variable to find its neighbors to
be connected in the Gaussian Markov random field defined by $\Theta$,
we call this result as neighborhood FDR control.

\section{Numerical Results}

We show that the theoretical properties discussed above also works in
simulated settings. 

\subsection{Quality of Estimation}
For all numerical examples here, we generate
$n=100,200,300,400$ random i.i.d. samples from
$\cN(0, \Sigma_{p\times p})$, where $p=500$ is fixed. We plant a
simple block diagonal structure into the true matrix $\Sigma$, which
is also preserved in the precision matrix $\Theta=\Sigma^{-1}$. All
the blocks had the same size of $4$, so that we have total $125$
blocks at the diagonal of $\Theta$. We set $\Theta_{ii}=1$ and set the
entries $\Theta_{ij}$, $i\neq j$, to a high enough value whenever
$(i,j)$ belongs to those blocks, in order to represent groups of
highly correlated variables. All experiments were repeated for $25$
times.

The $\lambda$ sequence for \gslope has been chosen according to
Section~\ref{sec:lambda} with respect to a target FDR level
$q=0.05$. For the \glasso, we have used the $\lambda$ value
discussed in Reference \cite{BanG08}, to control the family-wise error rate
(FWER, the chance of any false rejection of null hypotheses) to
$\alpha = 0.05$.

\subsubsection{Mean Square Estimation Error}

Recall our discussion of error bounds in Theorem~\ref{thm:pb.slope}
(off-diagonal scaled by $\hat\Theta_{ii}$) and
Theorem~\ref{thm:pb.diag} (diagonal), followed by
Section~\ref{sec:est.dis} where we roughly sketched the asymptotic
behaviors of $\hat\Theta_{ii}$ and~$\hat\Theta_{-i,i}$.

The top panel of Figure~\ref{fig:all} shows mean square error (MSE)
between estimated quantities and the true models that we have created,
where estimates are obtained by our method \gslope, without~or with
symmetrization at the end of Algorithm~\ref{alg:gslope}, as well as by
the \glasso~\cite{FriH08} which solves the $\ell_1$-based MLE
problem~\eqref{eq:glasso} with a block coordinate descent
strategy. The off-diagonal estimation was consistently good overall
settings, while the estimation error of diagonal entries were kept
improving as $n$ is being increased, which was we predicted in
Section~\ref{sec:est.dis}. We believe that our estimation of the diagonal
has room for improvement, for example, using more accurate reverse-scaling to
compensate the normalization within the SLOPE procedure.

\subsubsection{FDR Control}

A more exciting part of our results is the FDR control discussed in
Theorem~\ref{thm:fdr} and here we check how well the sparsity
structure of the precision matrix is recovered. For comparison, we measure
the power (the fraction of the true nonzero entries discovered by the
algorithm) and the FDR (for the whole precision matrix).

\begin{figure}[H]
  \centering
\includegraphics[width=.9\linewidth]{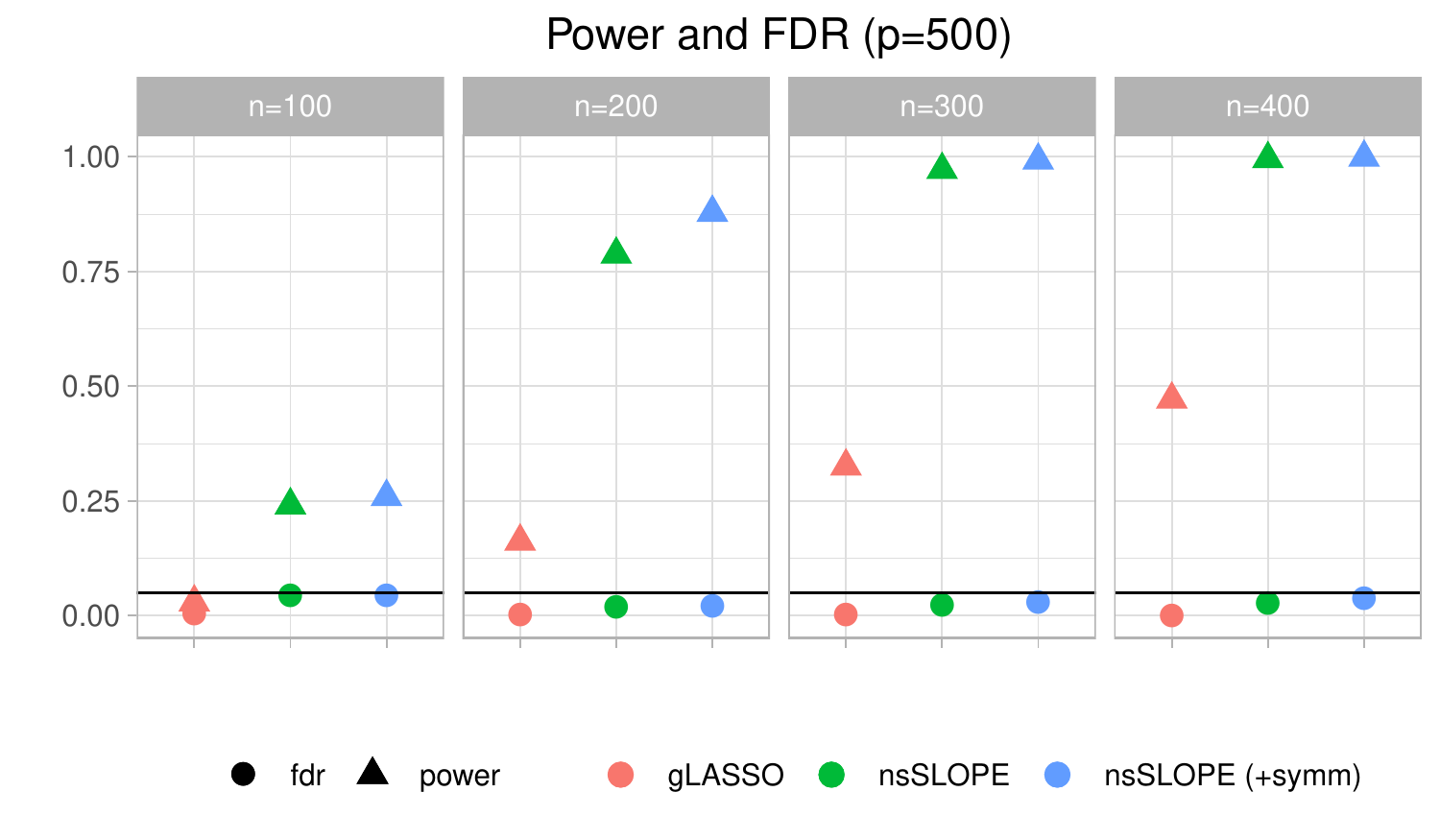} ~~~~
\includegraphics[width=.9\linewidth]{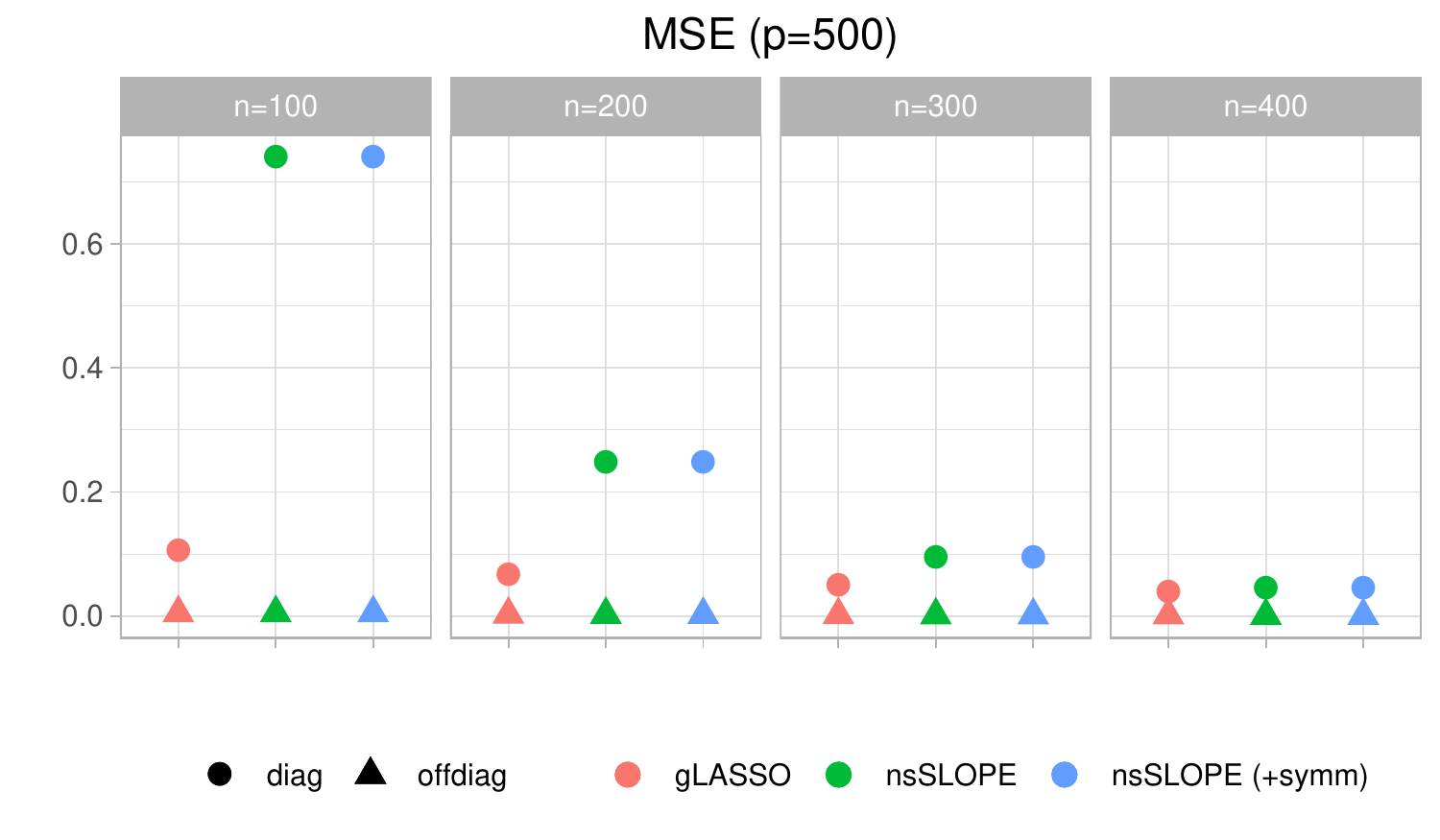}
\caption{Quality of estimation. Top: empirical false discovery rate (FDR) levels (averaged over 25 repetitions) and the nominal level of
  $q=0.05$ (solid black horizontal line). Bottom: mean square error of diagonal
  and off-diagonal entries of the precision matrix. 
    $p=500$ was fixed for both panels and
  $n=100$, $200$, $300$ and $400$ were tried. (``\gslope'': \gslope without symmetrization, ``+symm'': with symmetrization and \glasso.) \label{fig:all}}
\end{figure}

The bottom panel of Figure~\ref{fig:all} shows the statistics. In all
cases, the empirical FDR was controlled around the desired level
$0.05$ by all methods, although our method kept the level quite
strictly, having~significantly larger power than the \glasso. This is
understandable since FWER control of the \glasso if often too
respective, thereby limiting the power to detect true positive
entries. It is also consistent with the results reported for SL1-based
penalized regression~\cite{BogB15, SuC16}, which indeed is one of the
key benefits of SL1-based methods.

\subsection{Structure Discovery}

To further investigate the structure discovery by \gslope, we have experimented with two different types of covariance matrices—one with a block-diagonal structure and another with a hub structure. The covariance matrix of the block diagonal case has been constructed using the {data.simulation} function from the \texttt{varclust} R package, with $n=100$, $SNR=1$, $K=4$, $numb.vars=10$, $max.dim=3$ (this gives us a data matrix with $100$ examples and $40$ variables). In the hub structure case, we have created a $20$-dimensional covariance matrix with ones on the diagonal and $0.2$ in the first column and the last row of the matrix. Then we have used the {mvrnorm} function from the \texttt{MASS} R package to sample $500$ data points from a multivariate Gaussian distribution with zero mean and the constructed covariance matrix.


\begin{figure}[H]
  \centering
\includegraphics[width=.9\linewidth]{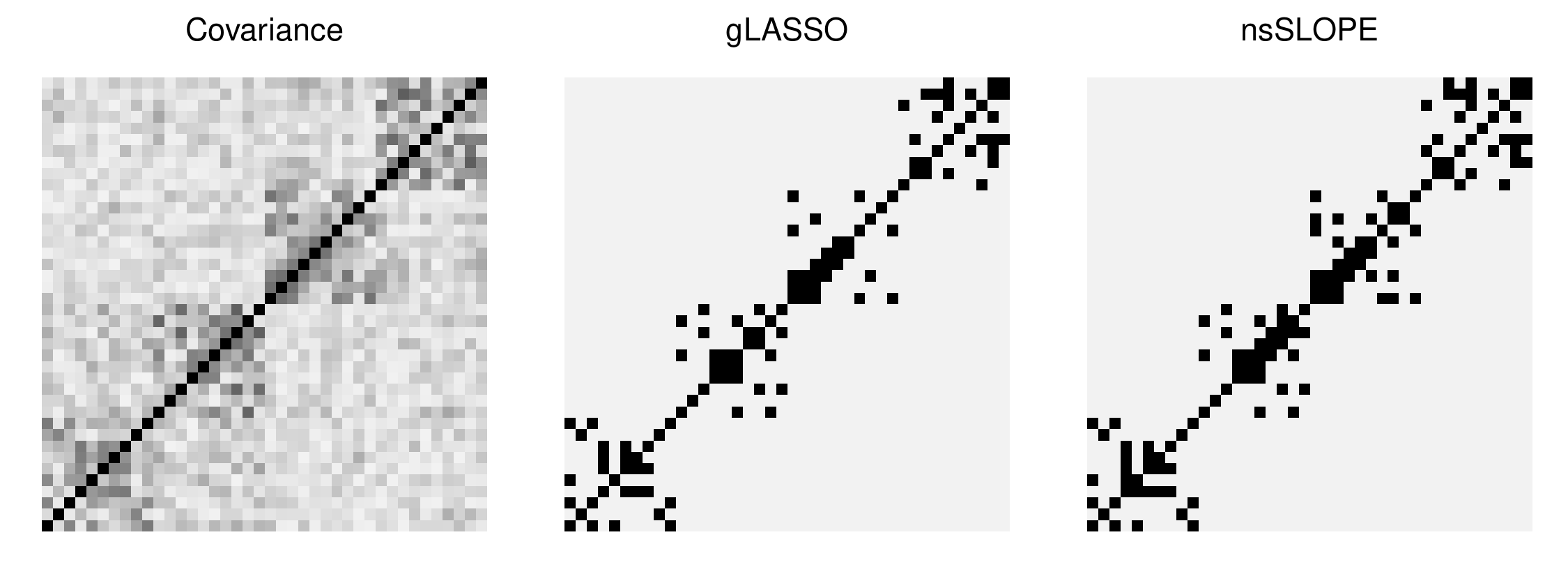} ~~~~
\includegraphics[width=.9\linewidth]{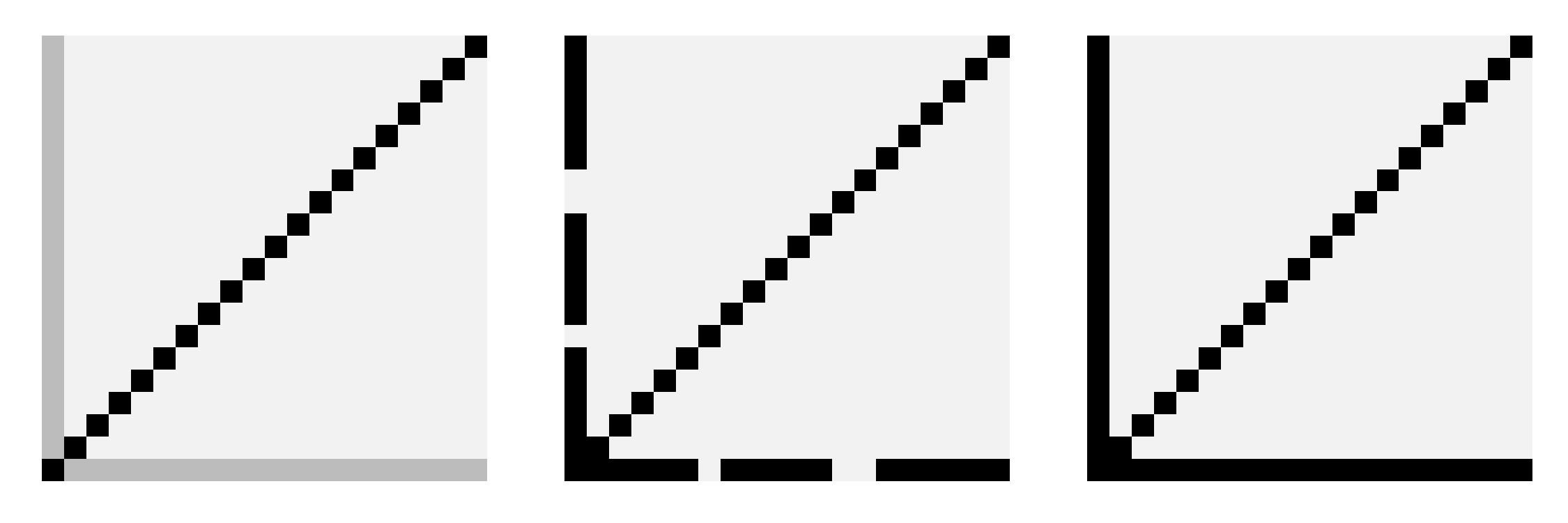}
\caption{Examples of structure discovery. Top: a covariance matrix with block diagonal structure. Bottom: a hub structure. True covariance matrix is shown on the left and \glasso and \gslope estimates (only the nonzero patterns) of the precision matrix are shown in the middle and in the right panels, respectively. \label{fig:struct}}
\end{figure}

The true covariance matrix and the two estimates from \glasso and \gslope are shown in Figure~\ref{fig:struct}. In \gslope, the FDR control has been used based on the Benjamini-Hochberg sequence~\eqref{eq:bh} with a target FDR value $q=0.05$ and in \glasso the FWER control~\cite{BanG08} has been used with a target value $\alpha=0.05$. Our method \gslope appears to be more sensitive for finding the true structure, although it may contain a slightly more false discovery.

\section{Conclusions}
We introduced a new procedure based on the recently proposed sorted
$\ell_1$ regularization, to find solutions of sparse precision matrix
estimation with more attractive statistical properties than the
existing $\ell_1$-based frameworks. We believe there are many aspects
of SL1 in graphical models to be investigated, especially when the
inverse covariance has a more complex structure. Still, we hope our
results will provide a basis for research and practical applications.

Our selection of the $\lambda$ values in this paper requires independence assumptions on features or blocks of features. Although some extensions are possible~\cite{BogB15}, it would be desirable to consider a general framework, for example, based on Bayesian inference considering the posterior distribution derived from the loss and the regularizer~\cite{ParC08,MalY14}, which enables us to evaluate the uncertainty of edge discovery and to find $\lambda$ values from data.

\vspace{6pt} 



\authorcontributions{Conceptualization, S.L. and P.S.; methodology, S.L. and P.S.; software, S.L. and P.S.; validation, S.L, P.S. and M.B.; formal analysis, S.L.; writing--original draft preparation, S.L.; writing--review and editing, S.L.}

\funding{This work was supported by the research fund of Hanyang University (HY-2018-N).}


\conflictsofinterest{The authors declare no conflict of interest.}


\reftitle{References}
\externalbibliography{yes}




\begin{thebibliography}{999}
\providecommand{\natexlab}[1]{#1}

\bibitem[Yuan and Lin(2007)]{YuaL07}
Yuan, M.; Lin, Y.
\newblock Model selection and estimation in the Gaussian graphical model.
\newblock {\em Biometrika} {\bf 2007}, {\em 94},~19--35.

\bibitem[{d'}Aspremont \em{et~al.}(2008){d'}Aspremont, Banerjee, and
  El~Ghaoui]{AspB08}
{D'}Aspremont, A.; Banerjee, O.; El~Ghaoui, L.
\newblock First-Order Methods for Sparse Covariance Selection.
\newblock {\em {SIAM} J. Matrix Anal. Appl.} {\bf 2008},
  {\em 30},~56--66.

\bibitem[Banerjee \em{et~al.}(2008)Banerjee, Ghaoui, and d'Aspremont]{BanG08}
Banerjee, O.; Ghaoui, L.E.; d'Aspremont, A.
\newblock Model Selection Through Sparse Maximum Likelihood Estimation for
  Multivariate Gaussian or Binary Data.
\newblock {\em J. Mach. Learn. Res.} {\bf 2008}, {\em
  9},~485--516.

\bibitem[Friedman \em{et~al.}(2008)Friedman, Hastie, and Tibshirani]{FriH08}
Friedman, J.; Hastie, T.; Tibshirani, R.
\newblock Sparse inverse covariance estimation with the graphical lasso.
\newblock {\em Biostatistics} {\bf 2008}, {\em 9},~432--441.

\bibitem[Tibshirani(1996)]{Tib96}
Tibshirani, R.
\newblock Regression Shrinkage and Selection via the Lasso.
\newblock {\em J. R. Stat. Soc. (Ser. B)} {\bf 1996},
  {\em 58},~267--288.

\bibitem[Oztoprak \em{et~al.}(2012)Oztoprak, Nocedal, Rennie, and
  Olsen]{OztN12}
Oztoprak, F.; Nocedal, J.; Rennie, S.; Olsen, P.A.
\newblock Newton-Like Methods for Sparse Inverse Covariance Estimation. In {\em
  Advances in Neural Information Processing Systems 25}; MIT Press:  Cambridge, MA, USA,  2012; pp.
  764--772.

\bibitem[Rolfs \em{et~al.}(2012)Rolfs, Rajaratnam, Guillot, Wong, and
  Maleki]{RolB12}
Rolfs, B.; Rajaratnam, B.; Guillot, D.; Wong, I.; Maleki, A.
\newblock Iterative Thresholding Algorithm for Sparse Inverse Covariance
  Estimation. In {\em Advances in Neural Information Processing Systems 25}; MIT Press:  Cambridge, MA, USA,  2012; pp. 1574--1582. 

\bibitem[Hsieh \em{et~al.}(2011)Hsieh, Dhillon, Ravikumar, and Sustik]{HsiD11}
Hsieh, C.J.; Dhillon, I.S.; Ravikumar, P.K.; Sustik, M.A.
\newblock Sparse Inverse Covariance Matrix Estimation Using Quadratic
  Approximation. In {\em Advances in Neural Information Processing Systems 24};
  MIT Press: Cambridge, MA, USA,  2011; pp.~2330--2338. 

\bibitem[Hsieh \em{et~al.}(2012)Hsieh, Banerjee, Dhillon, and
  Ravikumar]{HsiB12}
Hsieh, C.J.; Banerjee, A.; Dhillon, I.S.; Ravikumar, P.K.
\newblock A Divide-and-Conquer Method for Sparse Inverse Covariance Estimation.
  In {\em Advances in Neural Information Processing Systems 25}; MIT Press:  Cambridge, MA, USA,  2012; pp.
  2330--2338.

\bibitem[Hsieh \em{et~al.}(2013)Hsieh, Sustik, Dhillon, Ravikumar, and
  Poldrack]{HsiS13}
Hsieh, C.J.; Sustik, M.A.; Dhillon, I.; Ravikumar, P.; Poldrack, R.
\newblock {BIG} \& {QUIC}: Sparse Inverse Covariance Estimation for a Million
  Variables. In {\em Advances in Neural Information Processing Systems 26}; MIT
  Press: Cambridge, MA, USA,  2013; pp. 3165--3173. 

\bibitem[Mazumder and Hastie(2012)]{MazH12}
Mazumder, R.; Hastie, T.
\newblock Exact Covariance Thresholding into Connected Components for
  Large-scale Graphical Lasso.
\newblock {\em J. Mach. Learn. Res.} {\bf 2012}, {\em
  13},~781--794.

\bibitem[Treister and Turek(2014)]{TreT14}
Treister, E.; Turek, J.S.
\newblock A Block-Coordinate Descent Approach for Large-scale Sparse Inverse
  Covariance Estimation. In {\em Advances in Neural Information Processing
  Systems 27};  MIT Press:  Cambridge, MA, USA,   2014; pp. 927--935.

\bibitem[Zhang \em{et~al.}(2018)Zhang, Fattahi, and Sojoudi]{ZhaF18}
Zhang, R.; Fattahi, S.; Sojoudi, S.
\newblock\emph{ Large-Scale Sparse Inverse Covariance Estimation via Thresholding and
  Max-Det Matrix Completion}; 
\newblock  International Conference on Machine Learning, PMLR: Stockholm, Sweden, 2018.

\bibitem[Meinshausen and B\"uhlmann(2006)]{MeiB06}
Meinshausen, N.; B\"uhlmann, P.
\newblock High-dimensional graphs and variable selection with the Lasso.
\newblock {\em Ann. Stat.} {\bf 2006}, {\em 34},~1436--1462.

\bibitem[Meinshausen and B\"uhlmann(2010)]{MeiB10}
Meinshausen, N.; B\"uhlmann, P.
\newblock Stability selection.
\newblock {\em J. R. Stat. Soc. (Ser. B)} {\bf 2010},
  {\em 72},~417--473.

\bibitem[Rothman \em{et~al.}(2008)Rothman, Bickel, Levina, and Zhu]{RotB08}
Rothman, A.J.; Bickel, P.J.; Levina, E.; Zhu, J.
\newblock Sparse permutation invariant covariance estimation.
\newblock {\em \mbox{Electron. J. Stat.}} {\bf 2008}, {\em 2},~494--515.

\bibitem[Lam and Fan(2009)]{LamF09}
Lam, C.; Fan, J.
\newblock Sparsistency and rates of convergence in large covariance matrix
  estimation.
\newblock {\em  Ann. Stat.} {\bf 2009}, {\em 37},~4254--4278.

\bibitem[Raskutti \em{et~al.}(2009)Raskutti, Yu, Wainwright, and
  Ravikumar]{RasY09}
Raskutti, G.; Yu, B.; Wainwright, M.J.; Ravikumar, P.K.
\newblock Model Selection in Gaussian Graphical Models: High-Dimensional
  Consistency of $\ell_1$-regularized MLE. In {\em Advances in Neural
  Information Processing Systems 21};   MIT Press:  Cambridge, MA, USA, 2009; pp. 1329--1336. 

\bibitem[Yuan(2010)]{Yua10}
Yuan, M.
\newblock High Dimensional Inverse Covariance Matrix Estimation via Linear
  Programming.
\newblock {\em J. Mach. Learn.~Res.} {\bf 2010}, {\em
  11},~2261--2286.

\bibitem[Fattahi \em{et~al.}(2018)Fattahi, Zhang, and Sojoudi]{FatZ18}
Fattahi, S.; Zhang, R.Y.; Sojoudi, S.
\newblock Sparse Inverse Covariance Estimation for Chordal Structures.
\newblock  In~Proceedings of the 2018 European Control Conference (ECC), Limassol, Cyprus, 12--15 June {2018}; pp.~837--844. 
\bibitem[Bogdan \em{et~al.}(2015)Bogdan, van~den Berg, Sabatti, Su, and
  Candes]{BogB15}
Bogdan, M.; van~den Berg, E.; Sabatti, C.; Su, W.; Candes, E.J.
\newblock {SLOPE}---Adaptive Variable Selection via Convex Optimization.
\newblock {\em Ann. Appl. Stat.} {\bf 2015}, {\em 9},~1103--1140.

\bibitem[Brzyski \em{et~al.}(2015)Brzyski, Su, and Bogdan]{BrzS15}
\textls[10]{Brzyski, D.; Su, W.; Bogdan, M.}
\newblock \textls[10]{Group {SLOPE}---Adaptive selection of groups of predictors.}
\emph{arXiv} {\bf 2015}, {{arXiv:1511.09078}.}

\bibitem[Su and Cand{\`e}s(2016)]{SuC16}
Su, W.; Cand{\`e}s, E.
\newblock SLOPE is adaptive to unknown sparsity and asymptotically minimax.
\newblock {\em Ann. Stat.} {\bf 2016}, {\em 44},~1038--1068.

\bibitem[Bondell and Reich(2008)]{BonR08}
Bondell, H.D.; Reich, B.J.
\newblock Simultaneous Regression Shrinkage, Variable Selection, and Supervised
  Clustering of Predictors with {OSCAR}.
\newblock {\em Biometrics} {\bf 2008}, {\em 64},~115--123.

\bibitem[Figueiredo and Nowak(2016)]{FigN16}
Figueiredo, M.A.T.; Nowak, R.D.
\newblock Ordered Weighted {L1} Regularized Regression with Strongly Correlated
  Covariates: Theoretical Aspects.
\newblock  In Proceedings of the 19th International Conference on Artificial
  Intelligence and Statistics, {AISTATS} 2016, Cadiz, Spain, 9--11 May  2016; pp. 930--938.

\bibitem[Lee \em{et~al.}(2016)Lee, Brzyski, and Bogdan]{LeeB16}
Lee, S.; Brzyski, D.; Bogdan, M.
\newblock Fast Saddle-Point Algorithm for Generalized {Dantzig} Selector and
  {FDR} Control with the Ordered l1-Norm.
\newblock In  Proceedings of the 19th International Conference on Artificial
  Intelligence and Statistics (AISTATS),  Cadiz, Spain, 9--11 May 2016; Volume~51,  pp. 780--789. 

\bibitem[Chen and Banerjee(2016)]{CheB16}
Chen, S.; Banerjee, A.
\newblock Structured Matrix Recovery via the Generalized Dantzig Selector. In
  {\em Advances in Neural Information Processing Systems 29}; Lee, D.D., 
  Sugiyama, M., Luxburg, U.V., Guyon, I., Garnett, R., Eds.; Curran Associates,
  Inc.: Red Hook, NY, USA,  2016; pp. 3252--3260. 

\bibitem[Bellec \em{et~al.}(2017)Bellec, Lecu{\'e}, and Tsybakov]{BelLT17}
Bellec, P.C.; Lecu{\'e}, G.; Tsybakov, A.B.
\newblock Slope meets Lasso: Improved oracle bounds and optimality. \emph{arXiv} {\bf 2017}, {arXiv:1605.08651v3}.

\bibitem[Derumigny(2018)]{Der18}
Derumigny, A.
\newblock Improved bounds for Square-Root Lasso and Square-Root Slope.
\newblock {\em Electron. J. Stat.} {\bf 2018}, {\em
  12},~741--766.

\bibitem[Anderson(2003)]{And03}
Anderson, T.W.
\newblock {\em An Introduction to Multivariate Statistical Analysis};
  Wiley-Interscience: London, UK, 2003.

\bibitem[Beck and Tetruashvili(2013)]{BecT13}
Beck, A.; Tetruashvili, L.
\newblock On the Convergence of Block Coordinate Descent Type Methods.
\newblock {\em {SIAM} J. Optim.} {\bf 2013}, {\em
  23},~2037--2060.

\bibitem[Beck and Teboulle(2009)]{BecT09}
Beck, A.; Teboulle, M.
\newblock A Fast Iterative Shrinkage-Thresholding Algorithm for Linear Inverse
  Problems.
\newblock {\em SIAM J. Imaging Sci.} {\bf 2009}, {\em 2},~183--202.

\bibitem[Nesterov(1983)]{Nes83}
Nesterov, Y.
\newblock A Method of Solving a Convex Programming Problem with Convergence
  Rate $O(1/k^2)$.
\newblock {\em Soviet~Math. Dokl.} {\bf 1983}, {\em 27},~372--376.

\bibitem[Razaviyayn \em{et~al.}(2013)Razaviyayn, Hong, and Luo]{RazHL13}
Razaviyayn, M.; Hong, M.; Luo, Z.Q.
\newblock A Unified Convergence Analysis of Block Successive Minimization
  Methods for Nonsmooth Optimization.
\newblock {\em {SIAM} J. Optim.} {\bf 2013}, {\em
  23},~1126--1153.

\bibitem[Figueiredo and Nowak(2014)]{FigN14}
Figueiredo, M.; Nowak, R.
\newblock Sparse estimation with strongly correlated variables using ordered
  weighted $\ell_1$ regularization. \emph{arXiv} {\bf 2014}, 
\newblock {arXiv:1409.4005}.

\bibitem[Johnstone(2001)]{Joh01}
Johnstone, I.M.
\newblock Chi-square oracle inequalities.
\newblock {\em Lect. Notes-Monogr. Ser.} {\bf 2001}, {\em 36},~399--418.

\bibitem[Park and Casella(2008)]{ParC08}
Park, T.; Casella, G.
\newblock The Bayesian Lasso.
\newblock {\em J. Am. Stat. Assoc.} {\bf 2008},
  {\em 103},~681--686.

\bibitem[Mallick and Yi(2014)]{MalY14}
Mallick, H.; Yi, N.
\newblock A New Bayesian Lasso.
\newblock {\em Stat. Its Interface} {\bf 2014}, {\em 7},~571--582.

\end{thebibliography}
\end{document}